\def\year{2021}\relax
\documentclass[letterpaper]{article} % DO NOT CHANGE THIS
\usepackage{aaai21}  % DO NOT CHANGE THIS
\usepackage{times}  % DO NOT CHANGE THIS
\usepackage{helvet} % DO NOT CHANGE THIS
\usepackage{courier}  % DO NOT CHANGE THIS
\usepackage[hyphens]{url}  % DO NOT CHANGE THIS
\usepackage{graphicx} % DO NOT CHANGE THIS
\usepackage{graphicx}  % DO NOT CHANGE THIS
\usepackage{natbib}  % DO NOT CHANGE THIS AND DO NOT ADD ANY OPTIONS TO IT
\usepackage{caption} % DO NOT CHANGE THIS AND DO NOT ADD ANY OPTIONS TO IT

\usepackage{cite}

\usepackage{bm}

\usepackage{microtype}
\usepackage{subfigure}
\usepackage{booktabs} % for professional tables
\usepackage{multirow}
\usepackage{times}
\usepackage{natbib}

\usepackage{algorithm}
\usepackage{algorithmic}

\usepackage{amsmath}
\usepackage{amsfonts}
\usepackage{amsthm}
\usepackage{mathtools}

\usepackage{flushend}

\usepackage{booktabs}       %
\usepackage{amsfonts}       %
\usepackage{nicefrac}       %
\usepackage{microtype}      %

\usepackage{enumitem}
\setlist{nolistsep}
\setlist[itemize]{noitemsep, topsep=0pt}

\usepackage{times}

\usepackage{graphicx} %
\usepackage{color}

\usepackage{array}
\usepackage{xspace}
\usepackage{fancyhdr}
\usepackage{comment}

\usepackage[switch]{lineno}  %

\usepackage{amssymb}
\usepackage{bbding}
\usepackage{pifont}

\newcolumntype{H}{>{\setbox0=\hbox\bgroup}c<{\egroup}@{}}

\newcommand{\noaistats}[1]{}  %

\definecolor{darkgreen}{rgb}{0,0.4,0.0}
\definecolor{darkblue}{rgb}{0,0.1,0.3}
\definecolor{darkred}{rgb}{0.7,0.0,0.0}

\newcommand{\SUB}[1]{\ENSURE \hspace{-0.15in} \textbf{#1}}

\newcommand{\nc}{K}

\title{FDNAS: Improving Data Privacy and Model Diversity in AutoML}
\author{
    Chunhui Zhang$^1$, Yongyuan Liang$^2$, Xiaoming Yuan$^1$, and Lei Cheng$^3$\thanks{Corresponding author: Lei Cheng (leicheng@sribd.cn).} \\

}
\affiliations{
$^1$Northeastern University, $^2$Sun Yat-sen University, $^3$Shenzhen Research Institute of Big Data \\
chunhui.cheung@gmail.com, liangyy58@mail2.sysu.edu.cn, yuanxiaoming@neuq.edu.cn, leicheng@sribd.cn\\

}
\begin{document}
% \linenumbers  %
\maketitle

\begin{abstract}
\begin{quote}
To prevent the leakage of private information while enabling automated machine intelligence, there is an emerging trend to integrate federated learning and Neural Architecture Search (NAS). Although promising as it may seem, the coupling of difficulties from both two tenets makes the algorithm development quite challenging. In particular, how to efficiently search the optimal neural architecture directly from massive non-iid data of clients in a federated manner remains to be a hard nut to crack. To tackle this challenge, in this paper, by leveraging the advances in proxy-less NAS, we propose a Federated Direct Neural Architecture Search (FDNAS) framework that allows hardware-aware NAS from decentralized non-iid data of clients. To further adapt for various data distributions of clients, inspired by meta-learning, a cluster Federated Direct Neural Architecture Search (CFDNAS) framework is proposed to achieve client-aware NAS, in the sense that each client can learn a tailored deep learning model for its particular data distribution. Extensive experiments on real-world non-iid datasets show state-of-the-art accuracy-efficiency trade-offs for various hardware and data distributions of clients. Our codes will be released publicly upon paper acceptance. 

\end{quote}
\end{abstract}

\section{Introduction}
From AlexNet \citep{krizhevsky2012imagenet}, VGG \citep{simonyan2014very}, ResNet \citep{he2016deep}, to SENet \citep{hu2018squeeze}, tremendous research effort has been put in efficient deep learning model designs, leading to state-of-the-art performance for various machine learning tasks, including image recognition, object detection, and image segmentation \citep{ren2015faster, long2015fully}. The success of deep learning models relies on sufficient training on a vast amount of data with correct labels, which however are difficult to acquire in practice. In particular, the privacy issue, which has attracted much attention recently, makes data collection from clients much more challenging \citep{pmlr-v54-mcmahan17a}.

To train deep learning models adequately while protecting the privacies of clients, federated learning has come up as the major driving force to enhance the data security of deep learning methods \citep{pmlr-v54-mcmahan17a}. Instead of training models using collected data from clients, federated learning needs no data collection. Instead, each client trains the machine learning model on its local edge device and then uploads the model parameters to the central server. Since only machine learning models are exchanged in the air, the risk of the client's data leakage becomes much smaller \citep{truex2019hybrid}. Although straightforward as it may seem, integrating machine learning training methods into the framework of federated learning while with indistinguishable performance loss is of no simple matter. This has triggered recent theoretical research in both machine learning and optimizations \cite{li2020on}. 

Therefore, in federated learning, it is crucial to think about how to use automated machine learning to search for neural architectures directly from the clients’ data.
Thanks to gradient-based neural architecture search \citep{liu2018darts, cai2018proxylessnas}, the model search can be represented as updating architecture parameters. 
Some NAS methods search for backbone cells on proxy data to trim down computational cost \citep{liu2018darts, xu2020pcdarts}. 
However, these proxy strategies do not guarantee that the backbone cells have optimal performance on the target data\citep{yang2019evaluation}. 
More importantly, in federated learning, the difference in distribution between proxy data and target data will be larger due to the presence of non-iid data.
Therefore, there are enough reasons to propose a non-proxy, gradient-based Federated Direct Neural Architecture Search (\textbf{FDNAS}), with a feature of no data exchange. The first contribution of this paper is to propose such a scheme. 

On the other hand, due to prevalent human biases, preferences, habits, etc., clients are divided into different groups, in each of which the clients are similar to each other in terms of both data and hardware.
Therefore, instead of looking for an architecture that is too dense to suit each client, we expect that FDNAS should allow each client to use a lightweight architecture that fits their individual tasks.
To achieve this goal, We treat different clients' models as a large ensemble, in which the models are \textit{highly diverse and client-specific}.
Then, our \textbf{primary question} is: how can we effectively search for these clusters' diverse models that are near-optimal on their respective tasks in a federated manner, while their weights are entangled? To exploit the model diversity in such a complex ensemble, we resort to the fundamental idea of meta-learning. 

Particularly, in meta-learning, the model's weights that have been already meta-trained can be very efficient when they are adapted to different tasks via meta-test \citep{chen2019closerfewshot}.
Therefore, we propose to use the SuperNet in a meta-test-like manner in order to obtain all client-specific neural architectures in federated learning. In the order of ``meta-train" to ``meta-test", the SuperNet trained on all clients in the first phase are considered as the meta-training model for the next ``meta-test" client-specific adaptation. Following this idea, we propose the Cluster Federated Direct Neural Architecture Search (\textbf{CFDNAS}) that divides all clients into groups by data similarity, and each group is trained using the SuperNet from the previous phase. In turn, each group utilizes the previous SuperNet and can adjust the architecture to fit their own client data after only a few rounds of updates like a meta-test. Consequently, CFDNAS can quickly generate a specific architecture for each client's data in parallel, under the framework of federated learning.
The contributions of our FDNAS are summarized below: 
\begin{itemize}
\item[1.] 
We have integrated federated learning with gradient-based and proxy-less NAS. This allows federated learning not only to train weights but also to search model architectures directly from the clients' data.
\item[2.] 
Inspired by meta-learning, we extended FDNAS to CFDNAS to exploit the model diversity so that client-specific models can be quickly searched at very low computational cost.
\end{itemize}

\section{Related Work}
\textbf{Efficient neural architecture} designing is important for the practical deployment.
MobileNetV2 \citep{sandler2018mobilenetv2} proposes MBconv blocks, which largely reduce the model's FLOPs. Besides, efficient neural architecture search has recently gained increasing attention.
To speed up the NAS, the one-shot NAS approaches treat all normal nets as different subnets of the SuperNet and share weights among the operation candidates \citep{brock2018smash}.
ENAS \citep{pham2018efficient} uses the RNN controller to sample subnets in SuperNet and uses reinforce method to obtain approximate gradients of architecture.
DARTS \citep{liu2018darts} improves the search efficiency by representing each edge as a mixture of candidate operations and optimizing the weights of operation candidates in continuous relaxations.
Recently, hardware-aware NAS methods like ProxylessNAS, incorporate the latency feedback in search as the joint optimization task without any expensive manual attempts. \citep{wu2019fbnet, wan2020fbnetv2, cai2018proxylessnas}.
At the same time, some NAS methods search for the best backbone cells and transfer them to other target tasks by stacking them layer by layer \citep{liu2018darts, xu2020pcdarts}. 
However, there is a big difference between the proxy and the target here: typically, the distribution of proxy data differs from the target data, and the best block searched by the proxy method differs from the optimal normal net after stacked \citep{ yang2019evaluation}.
Their motivations and approaches are very different from our FDNAS, where our goal is to search the neural architecture without the gap between proxy task and clients' data, but directly on clients' data in a privacy-preserving way and automatically design a variety of client-specific models that satisfy clients' diversity.

\noindent \textbf{Federated learning} commonly deploys predefined neural architectures on the client. They then use FedAvg as a generic algorithm to optimize the model \citep{pmlr-v54-mcmahan17a}.
There are some fundamental challenges associated with the research topic of federated learning \citep{li2019federated}: communication overhead, statistical heterogeneity of data(non-iid), client privacy.
The communication between the central server and the client is a bottleneck due to frequent weights exchange. So some studies aim to design more efficient communication strategies \citep{konevcny2016federated, 45672, fedpaq19}.
In practice, the data is usually non-iid distributed. So the hyper-parameter settings of FedAvg (e.g., learning rate decay) are analyzed to study their impacts on non-iid data \citep{li2020on}. In addition, a global data-sharing strategy is proposed to improve the accuracy of the algorithm on non-iid data \citep{zhao2018federated}.
Other research efforts have focused on privacy security \citep{agarwal2018cpsgd}. For example, differential privacy is applied to federated training, thus preserving the privacy of the client \citep{article17eth, 9069945}. Recently, FedNAS searches out the cell architecture and stacks it \citep{he2020fednas}.
Note that all of the above techniques are deploying pre-defined network architectures, or only searching for backbone cells for stacking using proxy strategies. However, our FDNAS can search the complete model architecture directly from clients' data and allow the model to better adapt to the clients' data distribution while protecting privacy. Also, with CFDNAS, it is able to provide multiple suitable networks for diverse clients at a very low computational cost.
\section{Preliminary}
To motivate the idea of this paper, we briefly review the basics of federated learning and ProxylessNAS in this section. 
\subsection{Federated Learning}
\label{sec:FedAvg}

As an emerging privacy-preserving technology, federated learning (FedAvg) enables edge devices to collaboratively train a shared global model without uploading their private data to a central server \citep{pmlr-v54-mcmahan17a}. In particular, in the training round $t+1$, an edge device  $k\in S$ downloads the shared machine learning model  $\mathbf w^{k}_{t}$ (e.g., a CNN model) from the central server, and utilizes its local data to update the model parameters. Then, each edge device sends its updated model $\{\mathbf w^{k}_{t} \}_{k \in S}$ to the central server for aggregation. 

\subsection{ProxylessNAS}
\label{sec:plnas}
% The surging interest in NAS has been recently prompted by its great success in automating neural network design, which has outperformed many manually designed counterparts in various deep learning tasks.  However, most NAS methods need prohibitively intensive computational resources to achieve the best performance, making them still distant from being widely adopted in practice. To overcome this hurdle, t
% The framework of ProxylessNAS emerged recently that can save computational resources significantly\citep{cai2018proxylessnas}.  As a result, it opens the door to directly learn the optimal architecture from large-scale datasets, without resorting to a proxy-based scheme. Furthermore, it allows the incorporation of hardware latency into the neural architecture design for accelerating the inference on devices. %In this subsection, we briefly review its key idea and mechanisms. 

In ProxylessNAS, a SuperNet (over-parameterized net) is firstly constructed, which is denoted by a directed acyclic graph (DAG) with $N$ nodes. Each node $x^{(i)}$ represents a latent representation (e.g., a feature map), and each directed edge $e^{(i,j )}$ that connects node  $x^{(i)}$ and  $x^{(j)}$ defines the following operation:
\begin{align}
x^{(j)} =  \sum_{n=1}^N b_n^{(i \rightarrow j)} o_n \left[ x^{(i)} \rightarrow x^{(j)} \right],
\end{align}
where $o_n\left[x^{(i)} \rightarrow x^{(j)}\right] \in \mathcal O$ denotes a operation candidate (e.g., convolution, pooling, identity, ect.) that transforms $x^{(i)}$to $x^{(j)}$, and vector $\mathbf b^{(i \rightarrow j)} = [b_1^{(i \rightarrow j)}, \cdots, b_N^{(i \rightarrow j)}]$ is a binary gate that takes values as one hot vector which only set $b_n^{(i \rightarrow j)}=1$ with a probability $p_n^{(i \rightarrow j)}$ and other elements are 0 in forwarding.
% \begin{align}
% \mathbf b^{(i \rightarrow j)} =     \begin{cases}
%         [1, 0, \cdots, 0], & \text{with a probability $p_1^{(i \rightarrow j)}$}, \\
%         ~~~~~~~~\cdots \\
%         [0, 0, \cdots, 1], & \text{with a probability $p_N^{(i \rightarrow j)}$}. \\
%     \end{cases}
% \end{align}
Rather than computing all the operations in the operation set $\mathcal O$ \citep{liu2018darts}, there is only one operation $o_n \left[ x^{(i)} \rightarrow x^{(j)}\right]$ that is utilized to transform each node $x^{(i)}$ to one of its neighbors $x^{(j)}$ with a probability $p_n^{(i \rightarrow j)}$ at each run-time, thus greatly saving the memory and computations for training.
So it opens the door to directly learn the optimal architecture from large-scale datasets without resorting to proxy tasks. Furthermore, it applies the incorporation of hardware latency loss term into the NAS for reducing the inference latency on devices.

\section{Federated Neural Architecture Search}
\subsection{Motivation and Problem Formulation}
In the last section, ProxylessNAS is introduced as an efficient framework of great value for real-world applications since it is more efficient to use models directly searched from diverse daily data.
However, the training of ProxylessNAS needs to collect all target data in advance, which inevitably limits its practical use due to the privacy concern of clients. In contrast, FedAvg trains a neural network without exchanging any data and thus can protect the data privacy of clients. However, it usually does not take the NAS into account, and therefore requires cumbersome manual architecture design (or hyperparameter tuning) for the best performance. 

To integrate the merits from ProxylessNAS and FedAvg while overcoming their shortcomings, we develop a federated ProxylessNAS algorithm in this section. In particular, we propose a novel problem formulation as follows:
\begin{align}
    \min_{\pmb {\alpha}} \quad & \sum_{k=1}^K\mathcal{L}_{val}^{k}(\mathbf w^*(\pmb {\alpha}), \pmb {\alpha}) \label{eq:outerfednasFormulation}\\
    \text{s.t.} \quad &\mathbf w^*(\pmb {\alpha}) = \mathrm{argmin}_{\mathbf w} \enskip \sum_{k=1}^K\mathcal{L}_{train}^{k}(\mathbf w, \pmb {\alpha}), \label{eq:innerfednasFormulation}
\end{align}
where $\mathcal{L}_{val}^{k}(\cdot, \cdot)$ denotes the validation loss function of client $k$,   and $\mathcal{L}_{train}^{k}(\cdot, \cdot)$ denotes its training loss function. Vector $\pmb {\alpha}$ collects all architecture parameters $\{\alpha_n^{i \rightarrow j}, \forall i \rightarrow j, \forall n\}$ and  $\mathbf w$ collects all weights $\{\mathbf w^{i\rightarrow j}, \forall i\rightarrow j\}$.  This nested bilevel optimization formulation is inspired by those proposed in ProxylessNAS \citep{cai2018proxylessnas} and FedAvg \citep{pmlr-v54-mcmahan17a}. More specifically, for each client, the goal is to search the optimal architecture $\pmb {\alpha}$ that gives the best performance on its local validation dataset, while with the optimal model weights $\mathbf w^*(\pmb {\alpha})$ learnt from its local training dataset. %We refer to original paper for the more technical details \citep{cai2018proxylessnas}.

%
%\begin{figure}[t]
%    % \vspace{-10pt}
%    \centering
%    \includegraphics[width=1\linewidth]{fig/nas.pdf}
%    %\vspace{-15pt}
%    \caption{Illustration of federated search}
%    \label{fig:draft_fig1_fednas}
%    %\vspace{-15pt}
%\end{figure}

\subsection{Federated Algorithm Development}
\begin{figure*}[htbp]
    \centering
    \includegraphics[width=0.825\linewidth]{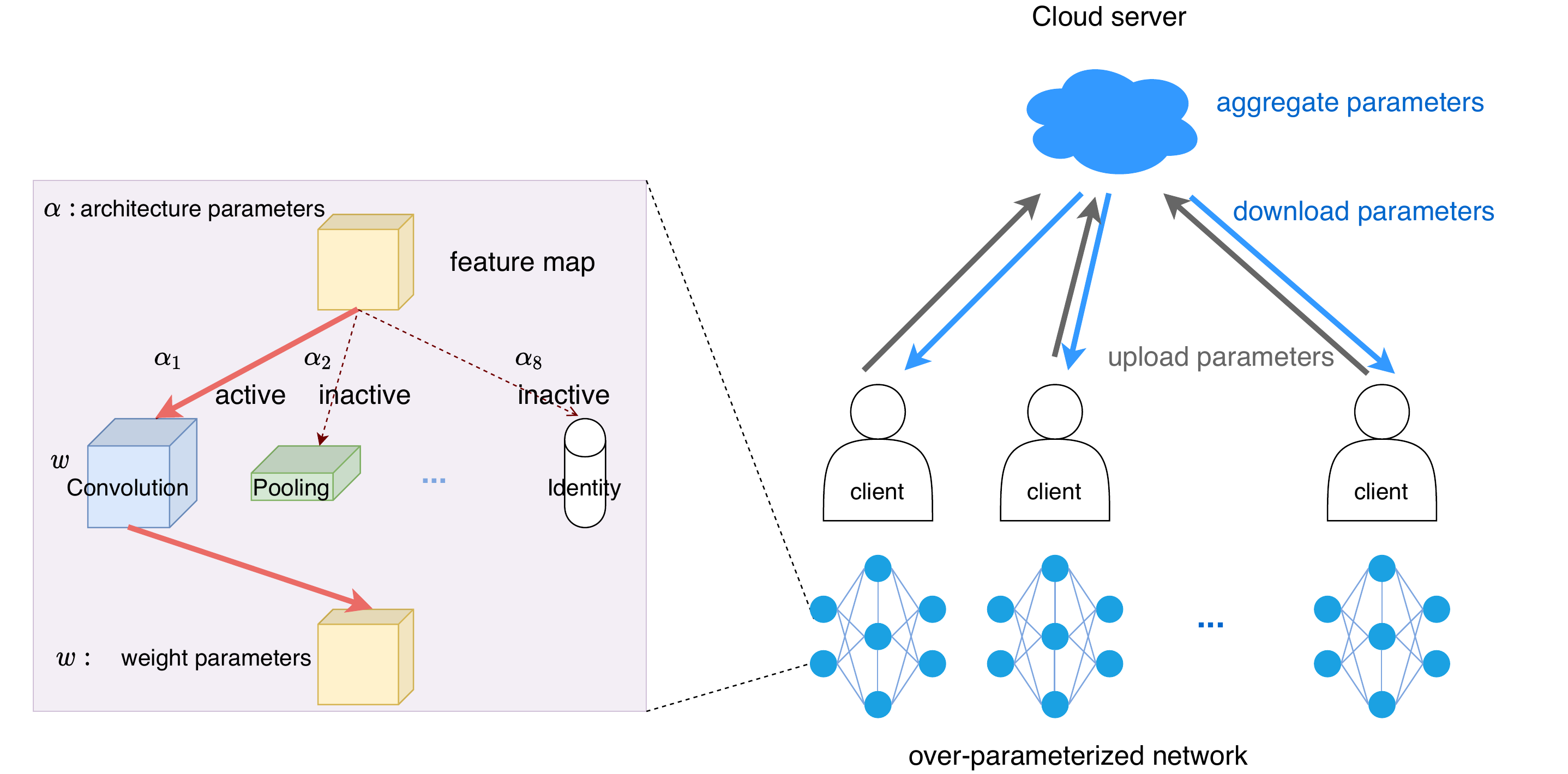}
    \caption{Illustration of FDNAS among clients.}
    \label{fig:fdnas}
\end{figure*}

To optimize $\pmb {\alpha}$ and $\mathbf w$ while protecting the privacy of client data, the FDNAS is developed in this subsection. 

Following the set-up of federated learning, assume that there are $K$ clients in the set $S$ and one central server. In communication round $t$, each client downloads global parameters  $\pmb {\alpha}^g_t$ and $\mathbf w^g_t$ from the server, and update these parameters using its local validation dataset and training dataset, respectively.  With global  parameters $\pmb {\alpha}^g_t$ and $\mathbf w^g_t$ as initial values, the updates follow the steps in {\bf Algorithm \ref{alg:fdnas}}, i.e., 
\begin{align}
    \mathbf w_{t+1}^k, \pmb \alpha_{t+1}^k \leftarrow \text{ProxylessNAS}( \mathbf w_t^{g}, \pmb \alpha_t^{g}), \forall k \in S. \label{eq:w-alpha-from-single-client}
\end{align}
Then, each edge device sends their updated parameters $\{\mathbf w^{k}_{t+1},\pmb \alpha^{k}_{t+1} \}_{k \in S}$ to the central server for aggregation. Central server aggregates these parameters to update global parameters $\pmb {\alpha}^g_{t+1}$ and $\mathbf w^g_{t+1}$:
\begin{align}
    \mathbf w_{t+1}^{g}, \pmb  \alpha_{t+1}^{g} \leftarrow \sum_{k=1}^\nc \frac{N_k}{N} \mathbf w_{t+1}^k, \sum_{k=1}^\nc \frac{N_k}{N} \pmb \alpha_{t+1}^k, \label{eq:sum-params-from-clients}
\end{align}
where $N_k$ is the size of local dataset in client $k$, and $N$ is the sum of all client's data size (i.e.,  $N = \sum_{k=1}^K N_k)$. 

After $T$ communication rounds, each client downloads the learnt global parameters $\pmb {\alpha}^g_t$ and $\mathbf w^g_t$ from the server, from which the optimized neural network architecture and model parameters can be obtained, as introduced in Section Preliminary. Furthermore, with the learnt neural network architecture, the weights $\mathbf w^k$ in each client could be further refined by the conventional FedAvg algorithm for better validation performance. The proposed algorithm is summarized in {\bf Algorithm \ref{alg:cfdnas}} and illustrated in Figure \ref{fig:fdnas}.

\begin{algorithm}[!h]
\caption{FDNAS: Federated Direct Neural Architecture Search}
\label{alg:fdnas}

\begin{algorithmic}
\SUB{Central server:}
%   \STATE {\bfseries search SuperNet}
   \STATE Initialize $\mathbf w_0^{g}$ and $\pmb \alpha_0^{g}$.
   \FOR{each communication round $t = 1, 2, \dots, T$}
    %  \STATE $m \leftarrow \max(\clientfrac\cdot K, 1)$
    %  \STATE $S_t \leftarrow$ (random set of $m$ clients)
     \FOR{each client $k \in S$ \textbf{in parallel}}
       \STATE $\mathbf w_{t+1}^k, \pmb \alpha_{t+1}^k \leftarrow \textbf{ClientUpdate}(\mathbf w_t^{g}, \pmb \alpha_t^{g})$
     \ENDFOR
     \STATE $\mathbf w_{t+1}^{g}, \pmb \alpha_{t+1}^{g} \leftarrow \sum_{k=1}^\nc \frac{N_k}{N} \mathbf w_{t+1}^k, \sum_{k=1}^\nc \frac{N_k}{N} \pmb \alpha_{t+1}^k$
    %  \STATE $\pmb \alpha_{t+1} \leftarrow \sum_{k=1}^\nc \frac{n_k}{n} \pmb \alpha_{t+1}^k$
   \ENDFOR
   \STATE
   
 \SUB{ClientUpdate($\mathbf w, \pmb \alpha$):} // \emph{On client platform.}
   \FOR{each local epoch $i$ from $1$ to $E$}
    \STATE update  $\mathbf w, \pmb \alpha$ by ProxylessNAS
  \ENDFOR
   \STATE return $\mathbf w, \pmb \alpha$ to server
\end{algorithmic}
\end{algorithm}

\subsection{Further Improvement and Insights}
\subsubsection{Clustering-aided model compression}
\begin{algorithm}[tb]
\caption{CFDNAS: Cluster Federated Direct Neural Architecture Search}
\label{alg:cfdnas}

\begin{algorithmic}
\SUB{Cluster server:}
   \STATE load $\mathbf w_0^{g}, \pmb \alpha_0^{g}$ by central server.
   \STATE $\{S_{1},\cdots,S_{P}\} \leftarrow$ (split $S_{all}$ into clusters by users' tag.)
   \FOR {cluster $S_{tag} \in \{S_{1},\cdots,S_{P}\}$ \textbf{in parallel}}
   \FOR{each round $t = 1, 2, \dots$}
    %  \STATE $m \leftarrow \max(\clientfrac\cdot K, 1)$
    %  \STATE $S_t \leftarrow$ (random set of $m$ clients)
     \FOR{each sampled client $k \in S_{tag}$ \textbf{in parallel}}
       \STATE $\mathbf w_{t+1}^k, \pmb \alpha_{t+1}^k \leftarrow \text{ClientUpdate}_k(\mathbf w_t^{g}, \pmb \alpha_t^{g})$ 
     \ENDFOR
     \STATE $\mathbf w_{t+1}^{g}, \pmb \alpha_{t+1}^{g} \leftarrow \sum_{k=1}^\nc \frac{n_k}{n_{all}} \mathbf w_{t+1}^k, \sum_{k=1}^\nc \frac{n_k}{n_{all}} \pmb \alpha_{t+1}^k$
   \ENDFOR
   \ENDFOR
   \STATE

\SUB{ClientUpdate($\mathbf w, \pmb \alpha$):} // \emph{On client platform.}
  \FOR{each local epoch $i$ from $1$ to $E$}
    \STATE update  $\mathbf w, \pmb \alpha$ by ProxylessNAS
 \ENDFOR
 \STATE return $\mathbf w, \pmb \alpha$ to server
\end{algorithmic}
\end{algorithm}

At the end of the proposed algorithm (i.e., {\bf Algorithm \ref{alg:cfdnas}}), the neural network architecture is learnt from the datasets of all the clients in a federated way. Although it enables the the knowledge transfer, accounting for every piece of information might result in structure redundancy for a particular group of clients. More specifically,  assume that 10 clients are collaborated together to train a model for image classification. Some clients are with images about \textit{birds, cats, and deer} and thus can form a ``animal" group, while other clients in the ``transportation" group are with images about \textit{airplanes, cars, ships, and trucks}. Although these images all contribute to the neural network structure learning in  {\bf Algorithm \ref{alg:cfdnas}}, the operations tailored for ``animal" group might not help the knowledge extraction from the ``transportation" group significantly. Therefore, an immediate idea is: could we further refine the model architecture by utilizing the clustering information of clients?

To achieve this, we further propose a clustering-aided refinement scheme into the proposed FDNAS. In particular, after $\pmb \alpha^{g}$converges, each client could send a tag about its data to the server, e.g.,  \textit{animal, transportation}, which are not sensitive about its privacy. Then the server gets the clustering information about the clients: which ones are with similar data distributions, based on which the clients' set $S$ can be divided into several groups:
\begin{align}
    S \rightarrow \{S_{1},\cdots,S_{P}\}.
    \label{eq:s-to-si}
\end{align}
Finally, the clients in the same group could further refine their SuperNet by re-executing the proposed FDNAS algorithm. The proposed clustering-aided refinement scheme labeled as \textbf{CFDNAS} is summarized in {\bf Algorithm \ref{alg:cfdnas}}.  

\subsubsection{Hardware-tailored model compression}
Clients might deploy models in different devices, such as mobile phone, CPU, and GPU. Previous studies show that taking hardware-ware loss into account could guide the NAS to the most efficient one in terms of the inference speed. 
For example, ProxylessNAS uses hardware-aware loss to \textit{avoid} $\pmb\alpha$ convergence to the heaviest operations at each layer.
The proposed FDNAS and CFDNAS could naturally integrate the hardware information into its search scheme. In particular, we could let each device send their hardware information such as \textit{GPU, CPU} to the server. Then, the server divides these clients into several groups according to their hardware type. For the clients in the same group, a hardware-ware loss term is added to the training loss. By taking this hardware-ware loss term into the training process, the proposed algorithm will drive the SuperNet to a compact one that gives the fastest inference speed for the particular hardware-platform. The details of hardware-ware loss term refer to ProxylessNAS.
%, which will be corroborated in the next section.

\begin{table*}[!htbp]
\centering
\begin{tabular}{lcccc}
\toprule
\textbf{\multirow{2}{*}{Architecture}} &\textbf{\multirow{2}{*}{Test Acc. (\%)}}   & \textbf{Params} & \textbf{Search Cost} & \textbf{\multirow{2}{*}{Method}} \\
\cmidrule(lr){3-4} &
 & \textbf{(M)} & \textbf{(GPU hours)} &\\
\hline
% ResNet-110~ \citep{he2016deep}                       & 93.57  & 1.7 & -    & manual \\
DenseNet-BC~ \citep{densenet}                       & 94.81  & 15.3 & -    & manual \\
MobileNetV2  ~ \citep{sandler2018mobilenetv2}                                   & 96.05  & 2.5   & -  & manual \\
\hline
NASNet-A ~ \citep{zoph2018learning}                 & 97.35      & 3.3  & 43.2K & RL      \\
AmoebaNet-A ~ \citep{amoebanet}           & 96.66      & 3.2  & 75.6K & evolution \\
% AmoebaNet-B ~ \citep{amoebanet}           & 97.45   & 2.8  & 75.6K & evolution \\
Hireachical Evolution~ \citep{liu2018hierarchical}          & 96.25   & 15.7 & 7.2K  & evolution \\
PNAS~ \citep{liu2018progressive}                            & 96.59    & 3.2  & 5.4K  & SMBO \\
ENAS~ \citep{pham2018efficient}                    & 97.11     & 4.6  & 12  & RL \\
\hline
DARTS~ \citep{liu2018darts}          & 97.24 & 3.3 & 72 & gradient \\
SNAS ~ \citep{xie2018snas}        & 97.02 & 2.9  & 36  & gradient \\
P-DARTS C10~ \citep{chen2019progressive}                                & 97.50 & 3.4  & 7.2 & gradient \\
P-DARTS C100~ \citep{chen2019progressive}                                & 97.38 & 3.6  & 7.2 & gradient \\
PC-DARTS ~ \cite{xu2020pcdarts} & 97.39 &3.6 &2.5 &  gradient\\
GDAS  ~ \citep{dong2019searching}                                   & 97.07 & 3.4  & 5 & gradient \\
% MobileNetV2  ~ \citep{sandler2018mobilenetv2}                                   & 96.05  & 2.5   & -  & manual \\
% MobileNetV2  ~ \citep{sandler2018mobilenetv2}                           & 72.12$^*$/95.85$\dag$ & 5.6  & - & manual \& federated \\
\hline
\textbf{FDNAS(ours)}                                  & \textbf{78.75$^*$/97.25$\dag$}  & \textbf{3.4}  & \textbf{59} & \textbf{gradient \& federated} \\
% central 96.03%, negligible
% \textbf{CFDNAS-G}                                   & -/\textbf{98.93$\dag$}  & \textbf{3.3}  & \textbf{3.5} & \textbf{gradient \& federated} \\
% \textbf{CFDNAS-C}                                   & -/\textbf{93.01$\dag$}  & \textbf{2.0}  & \textbf{3.5} & \textbf{gradient \& federated} \\
\bottomrule
\end{tabular}
\caption{\textbf{CIFAR-10 performance.} $^*$: The federated averaged model's accuracy. $\dag$: mean local accuracy, i.e., Each client completes training local epochs after downloading the server model, and then averages the local accuracy obtained by testing inference on the local test data.}
\label{tab_ev_cifar}
\end{table*}

\section{Experiments}

\subsection{Implementation Details}
We use PyTorch \citep{paszke2019pytorch} to implement FDNAS and CFDNAS.
We searched on CIFAR-10 \citep{cifar10} and then trained normal networks from scratch. CIFAR-10 has 50K training images, 10K test images, and 10 classes of labels. To simulate a federation scenario, we set up 10 clients. The first 3 classes of images are randomly and evenly assigned to the first 3 clients. Then, the middle 3 classes of images were randomly assigned to the middle 3 clients, and the last 4 classes of images were randomly and evenly assigned to the last 4 clients. 
Each client has 4500 images as a training set for learning $\mathbf w$ and 500 images as a validation set for learning $\pmb \alpha$.
Besides, to test the transferability of the architecture searched out by FDNAS, we used it to train on ImageNet\citep{5206848}.

\subsection{Image Classification on CIFAR-10}
\subsubsection{Training Settings}
% We use a total batch size of 256 and set the initial learning rate as $0.05$. Besides, the cosine rule is applied as a learning rate decay strategy. When training SuperNet, we use Adam \cite{adam14} to optimize $\pmb \alpha$ and the momentum SGD to optimize $\mathbf w$. 
We use a total batch size of 256 and set the initial learning rate to $0.05$. Then, we use the cosine rule as a decay strategy for the learning rate. When training SuperNet, we use Adam to optimize $\pmb \alpha$ and momentum SGD to optimize $\mathbf w$. The weight decay of $\mathbf w$ is $3e-4$, while we do not use weight decay on $\pmb \alpha$.
The SuperNet has 19 searchable layers, each consisting of MBconv blocks, the same as ProxylessNAS \citep{cai2018proxylessnas}.
We train the FDNAS SuperNet with 10 clients for a total of 125 rounds. Meanwhile, the local epoch of each client is 5. 
We assume that all clients can always be online during the training procedure.
After that, CFDNAS clusters clients-0, 1, and 2 into the GPU group to train CFDNAS-G SuperNet, and clients-3, 4, and 5 into the CPU group to train CFDNAS-C SuperNet. CFDNAS SuperNets are all searched separately for 25 rounds.
After training the SuperNet, we derive the normal net from the SuperNet and then run 250 rounds of scratch training on clients via FedAvg.
GPU latency is measured on a TITAN Xp GPU with a batch size of 128 in order to avoid severe underutilization of the GPU due to small batches. CPU latency is measured on two 2.20GHz Intel(R) Xeon(R) E5-2650 v4 servers with a batch size of 128.
Besides, the random number seeds were set to the same value for all experiments to ensure that the data allocation remained consistent for each training.

\subsubsection{Results}
% In centralized and gradient-based NAS, e.g., DARTS \citep{liu2018darts}, costs 72 GPU hours to obtain $97.24\%$ accuracy which spends 3 hours longer than our method. 
% Since our FDNAS needs no data collection to preserve privacy and works based on FedAvg, we use the federated averaged model's accuracy and clients' models mean local accuracy (clients train local epochs based on the downloaded server model and then test on the local test set) in Table~\ref{tab_ev_cifar}.
Since our FDNAS is based on FedAvg and protects the privacy of clients' data, we put the federated averaged model's accuracy and clients' models mean local accuracy in the Table~\ref{tab_ev_cifar}.
As demonstrated, our FDNAS normal net achieves $78.75\%$ federated averaged accuracy and $97.25\%$ mean local accuracy. 
It costs 59 GPU hours. However, the federated learning framework is naturally suited to distributed training. During training, all clients can be trained simultaneously. 
% Benefiting from the parallelism, it only takes $59/10$ hours to train 10 clients together and significantly reduces the actual training time for FDNAS in deployment.
Our FDNAS outperforms both evolution-based NAS and gradient-based DARTS in terms of search time cost, and our clients'local accuracy is higher than their central accuracy.
In addition, we use MobileNetV2 as a predefined, hand-crafted model trained under FedAvg for a fairer comparison with FDNAS. In Table~\ref{ablation1}, our FDNAS outperforms MobileNetV2 in terms of federated averaged accuracy and mean local accuracy. See ablation study~\ref{alation-sec} subsection for more analysis and CFDNAS's performance.

\subsection{Image Classification on ImageNet}
\begin{table*} [!t]
\centering
\begin{tabular}{lcccccc}
\toprule
\textbf{\multirow{2}{*}{Architecture}} & \multicolumn{2}{c}{\textbf{Test Acc. (\%)}} & \textbf{Params} & $\times+$ & \textbf{Search Cost} & \textbf{\multirow{2}{*}{Search Method}} \\
\cmidrule(lr){2-3}
& \textbf{top-1} & \textbf{top-5} & \textbf{(M)} & \textbf{(M)} & \textbf{(GPU hours)} &\\
\hline
MobileNet~ \citep{howard2017mobilenets}         & 70.6 & 89.5 & 4.2 & 569  & -    & manual \\
ShuffleNet 2$\times$ (v2)~ \citep{ma2018shufflenet}    & 74.9 & - & $\sim$5  & 591  & -    & manual \\
MobileNetV2~ \citep{sandler2018mobilenetv2}         & 72.0 & 90.4 & 3.4 & 300  & -    & manual \\
% MobileNetV2(1.4)~ \citep{sandler2018mobilenetv2}         & 74.7 & - & 6.9 & 585  & -    & manual \\

\hline
NASNet-A~ \citep{zoph2018learning}              & 74.0 & 91.6  & 5.3 & 564  & 43.2K & RL \\
NASNet-B~ \citep{zoph2018learning}              & 72.8 & 91.3  & 5.3 & 488  & 43.2K & RL \\
% NASNet-C~ \citep{zoph2018learning}              & 72.5 & 91.0  & 4.9 & 558  & 43.2K & RL \\
AmoebaNet-A~ \citep{amoebanet}        & 74.5 & 92.0  & 5.1 & 555  & 75.6K & evolution \\
AmoebaNet-B~ \citep{amoebanet}        & 74.0 & 91.5  & 5.3 & 555  & 75.6K & evolution \\
% AmoebaNet-C~ \citep{amoebanet}        & 75.7 & 92.4  & 6.4 & 570  & 75.6K & evolution \\
PNAS~ \citep{liu2018progressive}                & 74.2 & 91.9  & 5.1 & 588  & 5.4K  & SMBO \\
MnasNet~ \citep{tan2019mnasnet}              & 74.8 & 92.0  & 4.4 & 388  & -    & RL \\
\hline
DARTS ~ \citep{liu2018darts}      & 73.3 & 91.3  & 4.7 & 574  & 96    & gradient \\
SNAS ~ \citep{xie2018snas}     & 72.7 & 90.8  & 4.3 & 522  & 36  & gradient \\
ProxylessNAS ~ \citep{cai2018proxylessnas}       & 75.1 & 92.5  & 7.1 & 465  & 200  & gradient \\
P-DARTS-C10~ \citep{chen2019progressive}                 & 75.6 & 92.6  & 4.9 & 557  & 7.2  & gradient \\
P-DARTS-C100 ~ \citep{chen2019progressive}             & 75.3 & 92.5  & 5.1 & 577  & 7.2  & gradient \\
GDAS ~ \citep{dong2019searching}           & 74.0 & 91.5 & 5.3 & 581  & 5  & gradient \\\hline

\textbf{FDNAS(ours)}             & \textbf{75.3} & \textbf{92.9}  & \textbf{5.1} & \textbf{388}  & \textbf{59}  & \textbf{gradient~ \& federated} \\

\bottomrule
\end{tabular}
\caption{\textbf{ImageNet performance.} $\times+$ denotes the number of multiply-add operations(FLOPs).}
\label{ev_imagenet}
\end{table*}
\subsubsection{Training Settings}
To test the generality on larger image classification tasks, we moved the FDNAS general net to ImageNet for training. Following the general mobile setting \citep{liu2018darts}, we set the input image size to $224\times224$ and the model's FLOPs were constrained to below 600M.
We use an SGD optimizer with a momentum of 0.9. The initial learning rate is 0.4 and decays to 0 by the cosine decay rule. Then the weight decay is 4e-5. The dropout rate is 0.2.
In order to fit the FDNAS net to ImageNet's image size, Layer 1, 3, 6, 8, and 16 are set to the downsampling layers.

\subsubsection{Results}
As shown in Table~\ref{ev_imagenet}, we achieved SOTA performance compared to other methods. FDNAS test accuracy is $75.3\%$, which is better than GDAS, ProxylessNAS, SNAS, and AmoebaNet. Besides, our FLOPs are 388M, which is also smaller than them. The search cost is only 59 GPU hours, which is smaller than ProxylessNAS and makes sense in real-world deployments.
Also, for a fairer comparison, we compare FDNAS to MobileNetV2 since they are both composed of MBconv blocks.
Our FDNAS normal net's $75.3\%$ accuracy outperforms MobileNetV2 by $3.3\%$. Moreover, the MobileNetV2(1.4)'s FLOPs is 585M which is quite more dense than FDNAS 388M. but our FDNAS's accuracy still outperforms its $0.6\%$. 
% The accuracy of our FDNAS normal network is $75.3\%$, which is $3.3\%$ higher than MobileNetV2. In addition, MobileNetV2 (1.4) has FLOPs of 585M, which is a lot denser than FDNAS's 388M, but $0.6\%$ less accurate than our FDNAS. 
Summing up the above analysis, the model searched by our FDNAS in a privacy-preserving manner is highly transferable and it has an outstanding trade-off between accuracy and FLOPs. These show that learning the neural architecture from the data can do away with the bias caused by human effort and attain better efficiency.

\subsection{Ablation study}
\label{alation-sec}
\begin{table*}[!t]
%\begin{threeparttable}
\centering
\begin{tabular}{lcccccc}
\toprule
\textbf{\multirow{2}{*}{Architecture}} & \multicolumn{2}{c}{\textbf{Latency (ms})} & \textbf{Params}  & \textbf{$\times+$} & \textbf{Search Cost} & \textbf{\multirow{2}{*}{Test Acc.(\%)}} \\ 
\cmidrule(lr){2-3}
& \textbf{GPU} & \textbf{CPU} & \textbf{(M)} & \textbf{(M)} & \textbf{(GPU hours)} &\\ \hline
 MobileNetV2       & 52.31  &  890.69  &2.5 &296.5  & - &  68.45$\ddag$/96.51$\dag$\\ \hline

 FDNAS           & 52.78  &  600.17 &  3.4 &346.6  &  59.00 &  78.75$\ddag$/97.25$\dag$\\ \hline
 CFDNAS-G        & 40.33 & 463.86 &  3.3  &318.4 &  3.53 &  73.60$\ddag$/ 98.93$\dag$\\
 CFDNAS-C        & 31.00 & 186.52  &  2.0 &169.3 &  3.46 &  71.29$\ddag$/ 93.01$\dag$\\
\bottomrule
\end{tabular}
\caption{\textbf{A comparison between MobileNetV2, FDNAS and CFDNAS on CIFAR-10}: 
% $\times+$ denotes the FLOPs. 
$\dag$ and $\ddag$ is explained in Table~\ref{tab_ev_cifar}.}

% FDNAS denotes the model searched in all clients. 
% CFDNAS-G(C) denotes when the FDNAS is got from all clients, we then cluster all clients into several groups and keep on searching in groups.
\label{ablation1}
\end{table*}
\begin{table*}[!t]
\centering
\begin{tabular}{lccccccc}
\toprule
\textbf{\multirow{2}{*}{Architecture}} &\multirow{2}{*}{\textbf{Client ID}} & \textbf{Params}  & \textbf{$\times+$} &\textbf{Search Cost}  &{\textbf{Client Local}} &\textbf{Mean Local} \\ 
&       &                     \textbf{(M)} & \textbf{(M)} &\textbf{(GPU hours)}&\textbf{Acc.(\%)}   & \textbf{Acc.(\%)} &\\ \hline

FDNAS  % &client-0    &       &    && 98.14    &    \\ 
        &0,1,2    &  3.38     & 346.64   &59 & 98.14/99.35/98.90    & 98.79   \\ 
                        % &client-2    &       &    && 98.90    &    \\ 
naive-CFDNAS-G   %&client-0    &       &   &&97.56  &  \\
   &0,1,2    & 3.70      & 356.83   &18 &97.56/98.11/97.36  &97.67  \\ 
%   &client-2    &       &    &&97.36  &  \\ 
\textbf{CFDNAS-G}   %&client-0    &       &   &&\textbf{98.39}  &  \\
   &0,1,2    & \textbf{3.33}      & \textbf{318.44}   &\textbf{3.53} &\textbf{98.39/99.02/99.38}  &\textbf{98.93}  \\ \hline
%   &client-2    &       &    &&\textbf{99.38}  &  \\ \hline
FDNAS   %&client-3    &       &    && 94.21    &    \\ 
                        &3,4,5    &3.38 &346.64 &59 & 94.20/91.81/93.04    & 93.01   \\ 
                        % &client-5    &       &    & & 93.04    &    \\ \hline
naive-CFDNAS-C   %&client-3    &       &    &&88.61  &  \\
   &3,4,5    & 1.92     & 187.89   &18 &88.61/89.88/90.25  &89.58  \\ 
%   &client-5    &       &    && 90.25 &  \\ \hline
\textbf{CFDNAS-C}  % &client-3    &       &    &&\textbf{93.21}  &  \\
   &3,4,5    & \textbf{2.03}      & \textbf{169.35}   &\textbf{3.46} &\textbf{93.21/92.73/93.12}  &\textbf{93.02}  \\
%   &client-5    &       &    && \textbf{93.12} &  \\ 
\bottomrule

\end{tabular}
\caption{\textbf{Enhancement by CFDNAS}. ``naive-CFDNAS'' means that SuperNet for CFDNAS is not inherited from FDNAS, but is searched directly in the cluster group from scratch.}
\label{ablation3}
\end{table*}
\begin{table*}[!t]
\centering
\begin{tabular}{lccccc}
\toprule
\textbf{\multirow{2}{*}{Architecture}} & \textbf{Params}  & \textbf{$\times+$} &\textbf{Search Cost}  & \textbf{Test Acc.} \\ 
&   \textbf{(M)} & \textbf{(M)} &\textbf{(GPU hours)}& \textbf{(\%)}  \\ \hline
DNAS(on client-0)       & 2.54      & 264.35   &6.4 & 95.83        \\ 
DNAS(on client-1)       & 3.50      & 330.58   &6.5 & 97.04        \\ 
DNAS(on client-2)       & 2.85      & 269.76   &8.4 & 97.55        \\ 
DNAS(on client-3)       & 2.77      & 258.90   &9.9 & 87.33        \\ 
DNAS(on client-4)       & 2.28      & 233.99   &6.7 & 88.38        \\ 
DNAS(on client-5)       & 3.27      & 299.63   &7.5 & 87.53        \\ 
DNAS(on client-6)       & 2.78      & 297.03   &7.7 & 97.83        \\ 
DNAS(on client-7)       & 3.25      & 336.20   &7.1 & 96.83        \\ 
DNAS(on client-8)       & 2.68      & 263.89   &8.8 & 97.38        \\ 
DNAS(on client-9)       & 2.80      & 286.46   &7.8 & 97.47        \\ 
mean           & 2.87      & 284.08   &7.7 & 94.31        \\ \hline

DNAS(on collected CIFAR-10)           & 5.02      & 500.71   &24.5 & 96.71        \\ \hline
FDNAS(on all clients)          & 3.38      & 346.64   &59.0 & 78.75$\ddag$/97.25$\dag$\\ \hline
CFDNAS-G(on client-0, 1, 2)        & 3.33      & 318.44   &3.53 & 73.60$\ddag$/98.93$\dag$\\
CFDNAS-C(on client-3, 4, 5)       & 2.03      & 169.35   &3.46 & 71.29$\ddag$/93.01$\dag$\\
\bottomrule
\end{tabular}
\caption{\textbf{Compared with conventional DNAS.} $\dag$ and $\ddag$ is explained in Table~\ref{tab_ev_cifar}. The DNAS search algorithm is ProxylessNAS like FDNAS, but uses well-collected data rather than federated learning.
}
\label{ablation2}
\end{table*}

\subsubsection{Effectiveness of CFDNAS}
From the Table~\ref{ablation1}, we demonstrate CFDNAS for comparison.
CFDNAS-G (GPU platform) and CFDNAS-C (CPU platform) are trained for 25 epochs each based on the inherited FDNAS SuperNet.
Both CFDNAS normal nets have smaller FLOPs than FDNAS. Both nets also have lower GPU/CPU inference latencies than FDNAS and the hand-crafted MobileNetV2.
We then study the improvement in accuracy by CFDNAS in Table~\ref{ablation3}.
Benefiting from the clustering approach, for the GPU group (including clients-0, 1, and 2), our CFDNAS-G search model achieves $98.92\%$ accuracy. It is more accurate than the original FDNAS on clients of the same GPU group and requires only 3.53 GPU hours of SuperNet adaptation, which is a negligible additional cost.
The ``naive-CFDNAS'' has no inheritance parameters ($\mathbf w$ and $\pmb \alpha$) and no SuperNet's ``meta-test'' adaptation. As a result, the convergence time of naive-CFDNAS takes 18 GPU hours, which is 5.1 times that of CFDNAS. In the same GPU group, our CFDNAS-G is $1.26\%$ more accurate than naive-CFDNAS, while its FLOPs are smaller.
For the CPU group (including clients-3, 4, and 5), our CFDNAS-C also outperforms FDNAS and naive-CFDNAS-G in terms of accuracy and FLOPs.
CPU group's data is harder than others, but our CFDNAS-C is still more accurate than naive-CFDNAS-C and more stable than FDNAS.

Compared to FDNAS, CFDNAS gets a better trade-off between accuracy and latency due to the meta-learning mechanism.
Compared to naive-CFDNAS, CFDNAS gets better performance because it inherits $\mathbf w$ and $\pmb \alpha$ from the FDNAS SuperNet, which is fully trained with data from all clients.
Also, because the FDNAS-based ``meta-trained" SuperNet performs the search, it helps meta-adaptation cost less than naive-CFDNAS.
% In Table~\ref{ablation2}, local accuracy of client-3, 4, and 5 is significantly lower than others. Also, in Table~\ref{ablation3}, FDNAS's mean local accuracy on the CPU group is $93.01\%$ and lower than all clients' mean local accuracy $96.71\%$ in Table~\ref{ablation2}. 

\subsubsection{Contributions of federated mechanism}
In Table~\ref{ablation2}, we show the effects of using a traditional DNAS with collected data (including all data) and a single-client DNAS (including only local data, with no federated training).
DNAS searches the model directly on the clients' data but requires data collection in advance, and it achieves a central accuracy of $96.71\%$. However, our FDNAS still has higher local accuracy than DNAS while protecting data privacy.
Besides, our FDNAS has smaller FLOPs than single-client conventional DNAS results. Our federated averaged accuracy is $78.75\%$, and local accuracy is $97.25\%$ which is $2.94\%$ higher than single-client local average accuracy. Thanks to the federated mechanism, FDNAS can use data from a wide range of clients to search for more efficient models. At the same time, it trains models with higher accuracy. Privacy protection and efficiency will be beneficial for the social impacts and effectiveness of actual machine learning deployments.

\subsubsection{Contributions of directly search}
For a fairer comparison, we use MobileNetV2, which is also composed of MBconv blocks, as a predefined hand-crafted neural architecture trained in FedAvg.
We train the normal net of FDNAS with the collected data from all clients and obtain an accuracy of $96.03\%$ in a centralized way, a negligible difference compared to the centralized results of MobileNetV2.
Then we present the FedAvg results for FDNAS and MobileNetV2 in Table~\ref{ablation1}. 
Our FDNAS federated averaged accuracy and local accuracy are $6.63\%$ and $1.4\%$ higher than MobileNetV2, respectively.
Also, although both have similar GPU latency, FDNAS can be faster than MobileNetV2 on the CPU platform. So compared to the FDNAS, MobileNetV2 is not optimal for diverse clients in federated scenarios.
The FDNAS search architecture from the dataset is superior to hand-crafted models, and it demonstrates that the NAS approach can provide a significant improvement over human design in federated learning.
% benefit for practical impact: 1. acc 2. negligible additional cost
% enhancement
% highlight benefit:
% 1.cfdnas(light addtional cost and diversity)
% 2.fed mechanism
% 3.NAS mechanism

\section{Discussion and Future Work}
% As an extension to FDNAS, CFDNAS can search diverse models at very low additional computational cost to satisfy different types of clients.% Inspired by meta-learning, as an extension, CFDNAS can discover diverse models derived from FDNAS with high accuracy and speed, and at a very low additional computational cost.
% We propose FDNAS to search for models directly from clients' data while protecting their privacies.
% Our empirical results for FDNAS have achieved state-of-the-art accuracy-efficiency trade-offs compared to hand-crafted models, which also demonstrates that neural architectures directly searched from clients' data in federated learning are superior to predefined hand-crafted neural architectures. At the same time, our FDNAS model achieves state-of-the-art results on ImageNet in a mobile setting, which demonstrates the transferability of FDNAS.
% Inspired by meta-learning, CFDNAS, an extension to FDNAS, can discover diverse high-accuracy and low-latency models derived from SuperNet of FDNAS at a very low computational cost.
% In the future work, we will extend our FDNAS to search for different tasks such as object detection, semantic segmentation, model compression, etc., as well as larger-scale datasets with more categories. % we achieve great performance and efficiency improvement over predefined hand-crafted neural architecture in federated learning. 
% our FDNAS greatly improves the model's performance and efficiency compared to predefined hand-crafted neural architectures in federated learning.
We propose FDNAS, a privacy-preserving neural architecture search scheme that directly searches models from clients' data under the framework of federated learning. Different from previous federated learning approaches, the complete neural architecture is sought from clients' data automatically without any manual attempts. Extensive numerical results have shown that our FDNAS greatly improves the model's performance and efficiency compared to predefined hand-crafted neural architectures in federated learning.
On the other hand, our FDNAS model achieves state-of-the-art results on ImageNet in a mobile setting, which demonstrates the transferability of FDNAS. Moreover, inspired by meta-learning, CFDNAS, an extension to FDNAS, can discover diverse high-accuracy and low-latency models adapted from a SuperNet of FDNAS at a very low computational cost. In future work, we will extend our FDNAS to search for different tasks such as object detection, semantic segmentation, model compression, etc., as well as larger-scale datasets with more categories.
\newpage
\section{Ethical Impact}
AutoML has been widely used in many fields, such as searching deep neural networks and tuning hyper-parameters in computer vision and natural language processing. 
In general, NAS brings some important implications for the neural network design, and here we focus on the use of FDNAS to provide a NAS solution in a better  privacy-preserving way and the use of CFDNAS to provide a suitable neural network architecture for each client. There are many benefits of the schemes, such as exploiting the diversity of models, respecting the diversity of clients, and reducing risks of privacy and security. 

Nowadays, there is a lot of research about federated learning for improving deep learning's fairness, safety, and trustworthiness. 
% To reduce the reliance of federated learning on manual attempts, we believe that the impacts of FDNAS in real-world scenarios is worthy of attention.
To liberate expensive manual attempts in federated learning, we believe that the impacts of FDNAS in real-world scenarios are worthy of attention. In addition, privacy leak from the gradient update is still an open problem in machine learning. 
It is worth exploring a more secure AutoML system combining with some potential privacy protection methods such as differential privacy, homomorphic encryption, and secure multi-party computing.
% In order to liberate expensive manual attempts in joint learning, we believe that the impact of FDNAS in real-world scenarios deserves attention.

% \bibliographystyle{IEEEtran}
\bibliography{ref}
\end{document}